\documentclass[10pt,twocolumn,letterpaper]{article}

\usepackage{cvpr}              

\usepackage{graphicx}
\usepackage{amsmath}
\usepackage{amssymb}
\usepackage{booktabs}
\usepackage[ruled]{algorithm2e}
\usepackage{amsmath}
\usepackage{stfloats}
\usepackage{multirow}
\usepackage{float}
\usepackage{stfloats}
\usepackage{indentfirst}

\usepackage[pagebackref,breaklinks,colorlinks]{hyperref}

\usepackage[capitalize]{cleveref}
\crefname{section}{Sec.}{Secs.}
\Crefname{section}{Section}{Sections}
\Crefname{table}{Table}{Tables}
\crefname{table}{Tab.}{Tabs.}

\def\cvprPaperID{4} 
\def\confName{CVPR}
\def\confYear{2022}

\begin{document}

\title{Person Re-identification Method Based on Color Attack and Joint Defence}

\newcommand*{\affaddr}[1]{#1} 
\newcommand*{\affmark}[1][*]{\textsuperscript{#1}}
\newcommand*{\email}[1]{\texttt{#1}}

\author{%
	Yunpeng Gong\affmark[]\affmark[]\qquad Liqing Huang\affmark[]\thanks{Equal contribution}\qquad Lifei Chen\affmark[]\thanks{Corresponding author} \\
    \affaddr{\affmark[]College of Computer and Cyber Security, Fujian Normal University, P. R. China}\qquad
	\affaddr{\affmark[]}\\
	\email{\tt\small \affmark[]fmonkey625@gmail.com}\qquad
	\email{\tt\small \affmark[]\{lqhuang,clfei\}@fjnu.edu.cn}
}

\maketitle

\begin{abstract}
The main challenges of ReID is the intra-class variations caused by color deviation under different camera conditions. Simultaneously, we find that most of the existing adversarial metric attacks are realized by interfering with the color characteristics of the sample. Based on this observation, we first propose a local transformation attack (LTA) based on color variation. It uses more obvious color variation to randomly disturb the color of the retrieved image, rather than adding random noise. Experiments show that the performance of the proposed LTA method is better than the advanced attack methods. Furthermore, considering that the contour feature is the main factor of the robustness of adversarial training, and the color feature will directly affect the success rate of attack. Therefore, we further propose joint adversarial defense (JAD) method, which include proactive defense and passive defense. Proactive defense fuse multi-modality images to enhance the contour feature and color feature, and considers local homomorphic transformation to solve the over-fitting problem. Passive defense exploits the invariance of contour feature during image scaling to mitigate the adversarial disturbance on contour feature. Finally, a series of experimental results show that the proposed joint adversarial defense method is more competitive than a state-of-the-art methods.
\end{abstract}

\section{Introduction}
\label{sec:intro}
Person re-identification (ReID) is matching the same person across diferent cameras and scenes\cite{survey,market1501,baseline,Li_2021_CVPR}. This technology have been widely applied to video surveillance\cite{Hou_2021_CVPR,Tian_2021_CVPR,Liu_2021_CVPR}, image retrieval\cite{sketch-criminal,sketch-based}, criminal investigation\cite{sketch-criminal}, target tracking\cite{Beyer_2017_CVPR_Workshops} and others. ReID has been a challenging and hot problem since illumination, complex environment, occlusion, image blur and other factors. In recent years, many ReID works\cite{Hou_2021_CVPR,Tian_2021_CVPR,Liu_2021_CVPR,Beyer_2017_CVPR_Workshops,sketch-criminal,sketch-based} used deep-learning module, and have made great progress. However, Szegedy et al.\cite{Szegedy} found the deep-learning models are susceptible to attacks from adversarial samples, which will cause the network to completely change its prediction results. The works of~\cite{SongBai,Bouniot,Hongjun,Zhedong,Physical-World-Attacks} have proved that the ReID systems based on deep-learning have the same vulnerability. And these adversarial samples have only added a slight disturbance, which is hidden enough for the human visual system. It is very important to study the security of ReID systems because the insecurity may cause severe losses, eg., criminals may use adversarial disturbance to cheat the monitoring systems~\cite{Hongjun,Physical-World-Attacks}.

\begin{figure}[]
	\centering
	\includegraphics[width=1\linewidth]{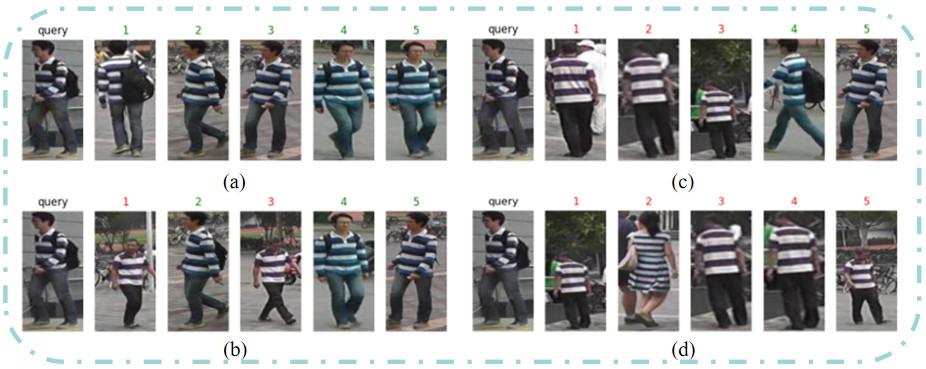}
	\caption{(a) shows the retrieval results of clean example. (b) corresponds to Meric-IFGSM attack~\cite{SongBai}, (c) corresponds to the SMA attack~\cite{Bouniot}, (d) corresponds to the proposed LTA attack. The numbers on the images indicate the rank of similarity in the retrieval results, the red and green number denote the wrong and correct results, respectively.}
	\label{fig:onecol}
\end{figure}

\begin{figure}[]
	\setlength{\abovecaptionskip}{0.1cm}
	\setlength{\belowcaptionskip}{-0.2cm}   
	\centering
	\includegraphics[width=1\linewidth]{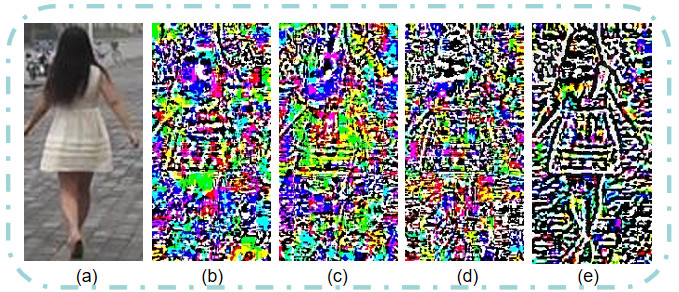}
	\caption{(a) shows clean example and (b) (c) (d) shows the adversarial noise generated after attacking three different models using SMA~\cite{Bouniot}, respectively. (b) corresponds to the normally trained model, (c) corresponds to the model which using~\cite{gyp} to train with better robustness to color variations, (d) corresponds to our proactive defense model, (e) corresponds to our joint adversarial defense.}
	\label{fig:onecol}
\end{figure}

The adversarial metric attack usually requires additional push or pull guidance to distort the distance between the attacked image and other images with the same identity or class~\cite{SongBai,Bouniot,Hongjun,ranking-attack-defense,flower-to-tower,Physical-World-Attacks,Relative-Order-Attack,black-box-attacks-retrieval,Universal-attack-retrieval,Enhancing-Adversarial-Robustness}, so as to achieve the purpose of deception models. Therefore, most of the existing studies~\cite{SongBai,Bouniot,ranking-attack-defense,	Enhancing-Adversarial-Robustness} of adversarial metric attack and defense revolve around metric relationships. Generally, the optimization function of the adversarial noise is designed for pulling the distance between negative pairs and pushing the distance between positive pairs. 

The intra-class variations caused by color deviations such as lighting, chromatic aberration, etc., in the various camera conditions is one of the main challenge for ReID~\cite{Joint-discriminative,Camera-style,Person-transfer-GAN}, due to the training set solely encompasses a limited portion of the intra-class variations of the color domain, the model is easy to overfit. In the existing adversarial metric attacks~\cite{SongBai,Bouniot,Hongjun}, it was observed that the attacks naturally perturb the color feature of the samples, which consistent with the color feature is important clue for image retrieval. The effect of the adversarial attack on the color feature is visualized in Figure $\color{red}1$. When retrieving clean samples, the model was able to identify the retrieved pedestrians with blue-gray striped top, grayish trousers, and backpack, even though they looked a little off from different cameras. While under adversarial perturbation, as shown in $\color{red}(b)$, $\color{red}(c)$, $\color{red}(d)$, the misjudgment of the model in color becomes more and more serious, which include the colors of pedestrian top, trousers and backpack. Two classic metric attacks include metric-IFGSM~\cite{SongBai} and SMA~\cite{Bouniot} attack. The metric-IFGSM~\cite{SongBai} attack was realized by maximizing the metric distance between the retreved image and the other images with the same identity, and used reference images. The SMA~\cite{Bouniot} attack added random noise to the retrieved image, and maximizing the metric distance from the clean image. The SMA not require reference images and thus is more with realistic scenario. Compared with metric-IFGSM, SMA only used the retrived image to generate disturbance, so it performs better in making the model misjudge color. Therefore, we further propose loacal transformation attack (LTA), which does not add random noise, but use local gray transformation with more obvious color variation to randomly disturb the color of the retrieved image, so as to learn robust adversarial noise against the color variation, and to strengthen the attack on color feature. Finally, experiments verify that the performance of the proposed LTA method is better than the advanced methods.

After exploring the vulnerability of ReID attack, we begin to research the effectiveness of defense methods.
Adversarial training is currently the main adversarial metric defense method~\cite{SongBai,Bouniot,ranking-attack-defense,Enhancing-Adversarial-Robustness}. Generally speaking, in adversarial training, a defense model trained by adversarial examples of an attack cannot defend against multiple attacks at the same time~\cite{Improving-Generalization-AT}, and extreme overfitting during training leads to obvious reduction in model generalization capacity~\cite{Robustness-Odds-with-Accuracy}.

Considering that being better at capturing shape or contour features is the main factor for the robustness of adversarial training~\cite{pmlr-v97-zhang19s}, and color features have a direct impact on the success rate of attacks. So there speculate that color features and contour feature are inherently important targets for attacker. To this end, we propose a corresponding joint adversarial defense approach. Firstly, we consider increasing the robustness of the model to color variations as a proactive defense. We speculate that when the robustness of the model to color variations is increased, the adversary will change the attack direction and strengthen the attack on the contour feature. It can be seen from Figure $\color{red}2(c)$ that the contour feature have been more seriously damaged. In addition, we also fuse sketch images during the model training process to strengthen the learning of contour feature so as to enhance the defense against the two attack modes (color and contour). From Figure $\color{red}2(d)$, it can be seen that the adversarial noise is significantly weakened on our proactive defense model. And then, we further propose a passive defense 
strategy, which utilizing  the invariance of contour features in the circuitous scaling to mitigate attack on contour feature. This strategically complementary to the proactive defense. From Figure $\color{red}2(e)$, it can be seen that after implementing the joint adversarial defense proposed in this paper, the adversarial noise becomes very sparse, and the contour feature are also well protected. The proposed joint defense model is a lightweight method without any additional parameter learning. It can be combined with various ReID models without changing the learning strategy. Therefore, the main contributions of this paper are summarized as follows:

$\bullet$ We propose a new attack method -- local transformation attack (LTA) for the first time, by using more obvious color variation to randomly disturb the color of the retrieved image and without reference image.

$\bullet$ We propose a joint adversarial defense model based on feature-invariant is to against adversarial metric attacks, which does not rely on adversarial training. The proposed method improves the robustness of the model, and performs well in cross-domain tests. 

$\bullet$ Finally, the comparative experimental results with the state-of-the-art algorithms further verified the effectiveness and the advanced nature of the proposed method.

\section{Related Work}
In this section, the previous work on adversarial attacks and defenses of the metric learning is described.
\label{sec:Related_Work}
\subsection{Adversarial Attacks}
Adversarial attacks can be categorized into white-box~\cite{white-box,SongBai} and black-box~\cite{black-box,Hongjun} attacks. The black-box attack means that the attacker does not know the structure and parameters of the target network, and the adversaries can only resort to the query access to generate adversarial samples. White-box attack assumes that the attacker has prior knowledge of the target networks, including the structure and parameters of model, which means that the adversarial examples are generated with and tested on the targe network. For the same attack, the success rate of white-box attack is higher than black-box attacks.

There are some metric attack methods proposed in ReID. Metric-FGSM~\cite{SongBai} extended some metric attacks by classification attacks, including fast gradient sign method (FGSM)~\cite{Goodfellow}, iterative FGSM (IFGSM) and momentum IFGSM (MIFGSM)~\cite{Yinpeng}. Among the three attack methods, IFGSM delivers the strongest white-box attacks~\cite{SongBai}. Opposite-direction feature attack (ODFA)~\cite{Zhedong} exploits feature-level adversarial gradients to generate adversarial examples to pull the feature in the opposite direction with an artificial guide. Self metric attack (SMA)~\cite{Bouniot} uses the image with added noise as the reference image and obtains the adversarial examples by attacking the feature distance between the original image and the reference image. This process does not require any additional images, it is more in line with the actual situation that the attacker usually lacks data. Furthest-negative attack (FNA)~\cite{Bouniot} combine hard sample mining~\cite{hard-mining,hard-mining2} and triple loss to obtain pushing guides and pulling guides to move image feature to head towards the least similar cluster of features while moving away from the other similar features. Deep mis-ranking (DMR)~\cite{Hongjun} proposed a learning-to-mis-rank formulation to perturb the ranking of the system output, which used a multi-stage network architecture that pyramids the features of different levels to extract general and transferable features for the adversarial perturbations. The success attack rate of black-box attacks is almost as high as that of white-box attack. At the same time, it also showed that when applied to classification attacks, it has a higher success attack rate than DeepFool~\cite{Deepfool}, NewtonFool~\cite{Uyeong}, and CW~\cite{Carlini}, and it has successfully broken through many classical ReID models\cite{p1,densenet,p2,Alignedreid,pcb,Li_2018_CVPR,duke,Zhong_2018_ECCV,Deng_2018_CVPR,Camera-style}.

\subsection{Adversarial Defenses}
Recently, a number of effective defense methods have been employed to against adversarial classification attacks~\cite{adversarial-transformations,JPG_compression,RRP,PGD,Denoising,Ground-Truth,adv_lib}, such as denoising methods, randomization-based schemes, adversarial training and others. The defense methods based on denoising, such as Guo et al.~\cite{adversarial-transformations} used more diversified non-differentiable image transformation operations, which include depth reduction, total variance minimization and image quilting. The goal is to increase the difficulty of network gradient prediction, and then achieve the purpose of defense. Noting that most of the training images are in JPG format, Dziugaite~\cite{JPG_compression} used JPG image compression method to reduce the impact of adversarial disturbance. In terms of randomization, RRP (random resizing and padding)~\cite{RRP} mitigates adversarial effects by combining random resizing and random padding based on adversarial training.~\cite{PGD,Denoising,Ground-Truth,adv_lib} showed that adversarial training is a robust way to defend against adversarial attacks, which includes offline adversarial training and online adversarial training. 

The metric defense schemes employ by~\cite{SongBai,Bouniot,Enhancing-Adversarial-Robustness,ranking-attack-defense} correspond to offline adversarial training and online adversarial training respectively. The defense method uses in~\cite{SongBai} is offline adversarial training, which is based on a generation of an adversarial version of the training set obtained with a frozen version of the trained model. As a frozen model is used to generate attacks, this method is referred to as offline adversarial training. The defense method uses in~\cite{Bouniot,Enhancing-Adversarial-Robustness,ranking-attack-defense} is online adversarial training, which generates adversarial examples online while the defended model evolves by triplet loss. However, adversarial training is prone to overfitting~\cite{Robustness-Odds-with-Accuracy,Bouniot} because dependent on the training data results in reducing the generalization capacity of the model. Enhancing the robustness towards adversarial examples and maintaining the generalization capacity of the model is the important issue of adversarial defense. 

\section{Proposed Methods}
\label{sec:}
In this section, we propose the local transformation attack (LTA) based on color features. In order to push the feature of the reference image away from the original image, there constructs a reference image with local difference from the original image in each basic iteration based on LGT~\cite{gyp}. As for the proposed attack method (LTA), we further propose a joint defense method. In proactive defense, we combine the three modal images of visible (RGB), grayscale and sketch for random channel fusion. In passive defense, it realizes by circuitous scaling of image. The specific attack and joint defende method framework is showed in Figure $\color{red}3$.	

\begin{figure*}[htbp] 
	\setlength{\abovecaptionskip}{0.1cm}
	\setlength{\belowcaptionskip}{-0.4cm}   
	\centering
	\includegraphics[width=1\linewidth]{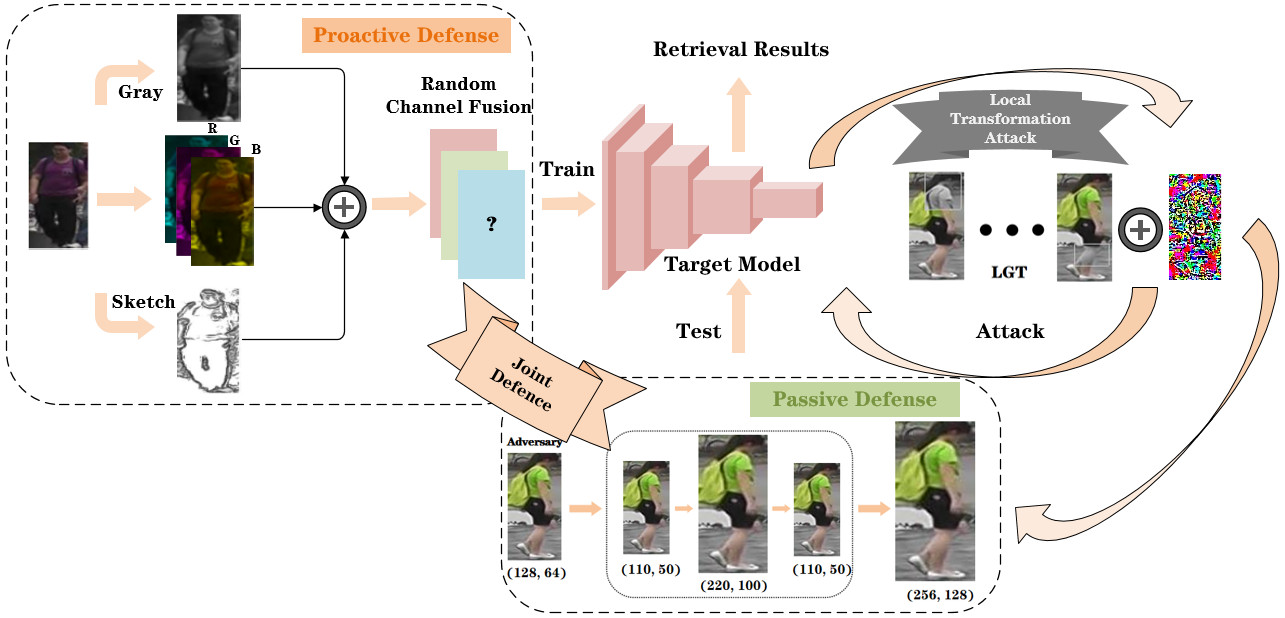}
	\caption{Framework diagram of our attack and joint adversarial defence. In our Local Transformation Attack (LTA), It pushes the feature of the reference image away from the original image by constructing a reference image with local difference from the original image in each basic iteration based on LGT~\cite{gyp}. In proactive defense, we combines the three modal images of visible (RGB), grayscale and sketch for random channel fusion. In our passive defense, it realizes by circuitous scaling of image.}
\end{figure*}
\subsection{Proposed Local Transformation Attack}
In order to attack the color feature, we propose the local transformation attack (LTA), which adopts local grayscale transformation (LGT)~\cite{gyp} constructing the local color deviation of the input. And then, it randomly selects a rectangular area in the image and replaces it with the pixels of the same rectangular area in the corresponding grayscale image. As showed in Figure $\color{red}3$, the LGT makes the constructed reference image have appropriate local differences from the original image. 

The initialization of the proposed LTA method is defined as:
\begin{equation}
	x_{adv}^{(0)} = x
\end{equation}
 where $x$ denote the attacked image. There using $x_{adv}^{(n)}$ denotes the adversarial example at the $n$-th iteration. $\hat{x}^{(n)}$ is the reference image with local variability constructed by LGT\cite{gyp} at the $n$-th basic iteration. So the proposed LTA is defined as the following iterative optimization:
\begin{equation}
	\hat{x}^{(n)} = LGT(x)
\end{equation}
\begin{equation}
	x_{adv}^{(n+1)} = \Psi_x^\varepsilon(x_{adv}^{(n)}+\alpha \cdot sign(grad^{(n+1)}))
\end{equation}
where $\epsilon$ is the adversarial bound and  $\alpha$ is the iteration step size, $\Psi_x^\varepsilon$ is the clip function, which ensures that $\|x_{adv}^{(n+1)} - x\|_\infty < \epsilon$ and that adversarial noise inconspicuousness. And the $grad^{(n)}$ is the accumulated gradient at the $n$-th iteration:
\begin{equation}
	grad^{(n+1)} = \theta \cdot grad^{(n)} + \frac{\Delta_{LTA}^{(n)}}{\|\Delta_{LTA}^{(n)}\|_1}
\end{equation}
where $\theta$ is the decay factor of the momentum term, in our experiments $\theta$ is set to 1. And  $\Delta_{LTA}^{(n)}$ is calculated as follows:
\begin{equation}
	\Delta_{LTA}^{(n)} = \frac{\partial D(	f_{adv}^{(n)}, 	\hat{f}^{(n)})}{\partial f_{adv}^{(n)}}
\end{equation}
\begin{equation}
	D(	f_{adv}^{(n)}, 	\hat{f}^{(n)}) = \|	f_{adv}^{(n)} - 	\hat{f}^{(n)}\|_2^2
\end{equation}
$f_{adv}^{(n)}$ denotes the feature of the adversarial example. Specifically, each iteration optimizes adversarial noise by attacking the feature distance between the adversarial image generated from the result of previous iteration and the new reference image.

\subsection{Proposed Joint Adversarial Defence Method}
In order to overcome the attack based on color features, we further propose the joint adversarial defense method (JAD). The proposed method includes the proactive and the passive defense. The proactive defense consists of channel fusion (CF) and local homogeneous transformation (LHT), and the passive defense consists of circuitous scaling (CS). 
\subsubsection{Proposed Proactive Defence}
The proactive defense consists of channel fusion (CF) and local homogeneous transformation (LHT).

\textbf{Channel fusion (CF)}. Visible images, grayscale images, and sketch images are homogeneous, which contain the same structural information. The results in~\cite{Homogeneous_Augmented,gyp} showed that using the homogeneous grayscale images to learn structural information in training is effective in increasing the robustness to color variations.

We add grayscale information and sketch information by channel fusing. The operation (Grayscale(3)) in Pytorch is adopted to get the grayscale image for each visible image, and sketch images can be obtained by inverting the grayscale image and then Gaussian blurring it, finally blending it with the grayscale image. As shown in Figure $\color{red}3$, the RGB images are randomly converted with a certain probability into 3-channel grayscale images or sketch images in pre-processing stage, and then randomly merge the channels of the grayscale image and the sketch image with the channels of the RGB image to create a new homogeneous modal image.

In the process of CF, 1 or 2 channels are randomly selected from the R, G, and B channels of the visible image. After the visible image channel and the number of channels n is determined, the grayscale or sketch image channel is randomly selected to reconstruct a new 3-channel image. In fact, a maximum of 60 homogeneous variations can be generated by combining the 5 image channels types of R, G, B, grayscale, and sketch in random order.

In addition, from Figure $\color{red}4a$, we can see that the augmentation based on image transformation will further extend the diversity of modes, such as Posterize, Equalize, Solarize, Contrast, Inversion and so on. However, the multi-modal inputs will lead to overfitting of the model, which affects the generalization capacity of the model as showed in Figure $\color{red}5a$.

\textbf{Local homogeneous transformation (LHT)}. To solve this problem, we propose a strategy method based on local homogeneous transformation (LHT), which extends local grayscale transformation (LGT)\cite{gyp}. Using LHT to guide the model to fit the diversity of variation gradually from local variations. The difference of LGT is that the proposed LHT replaces randomly selected regions with homogeneous images. As showed in Figure $\color{red}5b$, it positively helps reduce the overfitting in training. Unless otherwise specified, the diversity data learning in subsequent experiments combine with LHT by default.

\begin{figure}[t]
	\centering
	\includegraphics[width=0.8\linewidth]{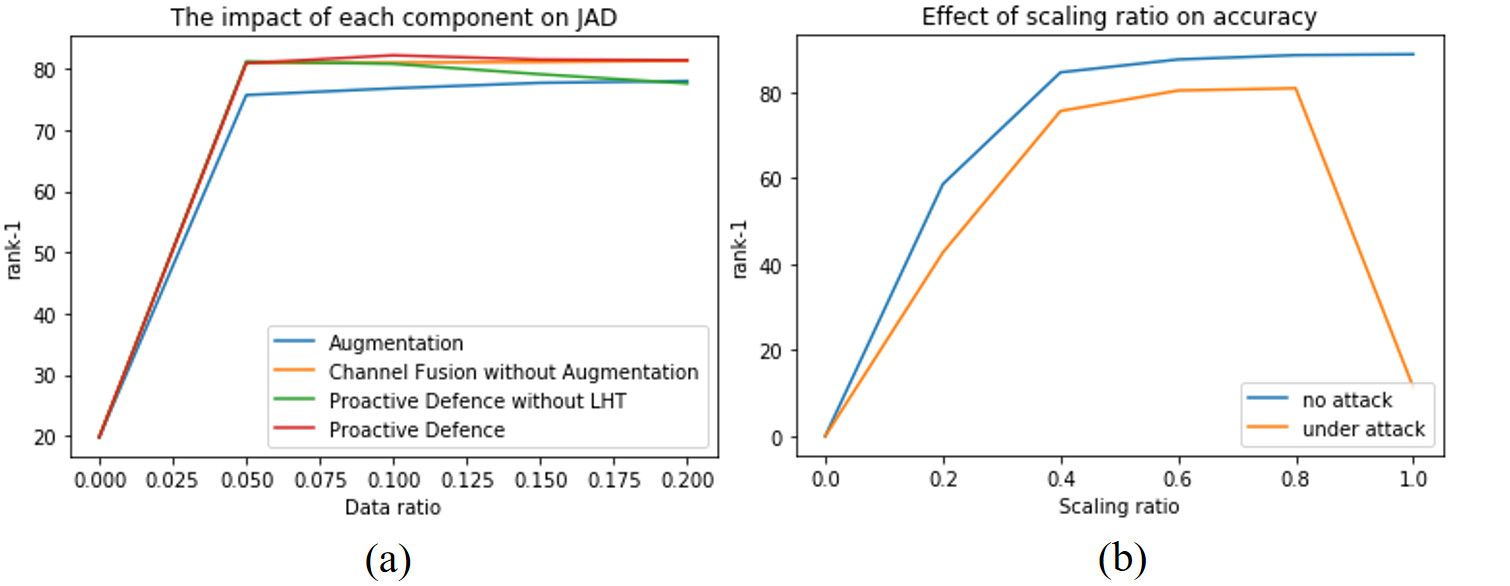}
	\caption{Hyper-parameter sensitivity analysis: (a) the contribution of each component of the joint defense to the defense and the impact of different ratios on defense performance; (b) the effect of different image scaling ratios on defense performance in passive defense.}
\end{figure}

The LHT for each visible image $x^v$ can be achieved by the following equation:

\begin{equation}
	x^h = T(x^v)
\end{equation}
and
\begin{equation}
	rect = RandPosition(x^v)
\end{equation}
and
\begin{equation}
	x^{LHT} = LT(x^v,x^h,rect)
\end{equation}
and
\begin{equation}
	(x^{LHT}|y) = (x^v|y)
\end{equation}
where the $x^h$ is the homogeneous images, and $T(\bullet)$ is the homogeneous transformation funtion. $RandPosition(\bullet)$ is used to generate a random rectangle in the image, and the function of $LT(\bullet)$ is to give the pixels in the rectangle corresponding to the $x^h$ image to the $x^v$ image. $x^{LHT}$ is the sample after local homogeneous transformation, and $y$ is the label of the transformed image.
\begin{figure}[t]
	\setlength{\abovecaptionskip}{0.1cm}
	\setlength{\belowcaptionskip}{-0.4cm}   
	\centering
	\includegraphics[width=0.8\linewidth]{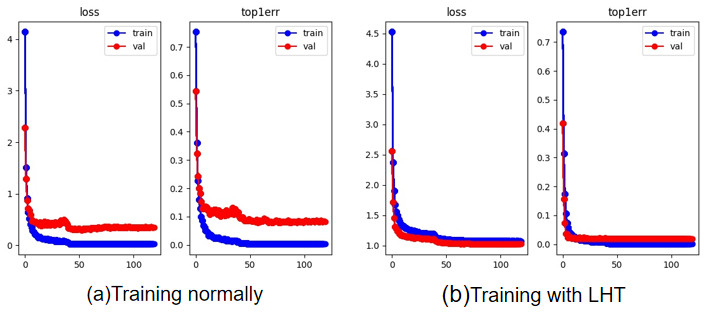}
	\caption{The comparison of training curve without LHT and with LHT.}
\end{figure}

From Figure $\color{red}4a$, we can see that the best defensive performance is achieved when the components are in the ratio of 5\% to 15\%. Therefore, the probability of the image using augmentation transformation is set to 5\%, and the probability of converting to a grayscale image is set to 5\%. In addition, the probability of using CF transformation is set to 5\%, and the probability of using LHT is set to 10\%.


\subsubsection{Proposed Passive Defence}
Since color features and contour feature are two important targets in the attack, the increased robustness of the model to color variations will force the adversary to change the direction of the attack to some extent, more towards attacking contour feature. Therefore, we exploit the invariance of contour features during image scaling to mitigate the adversarial disturbance on contour feature.

The basic principle of image scaling is to calculate the pixel value of the target image according to the pixel value of the original image by certain rules, common image scaling algorithms such as linear interpolation \cite{linear-interpolation}. In the scaling process, some pixels are discarded or some new pixels are added. \cite{RRP} found that the adversarial noise structure can be effectively destroyed by one-time image scaling. Circuitous scaling (CS) consists of multiple image scaling to give full play to this advantage.

The passive defense is realizes by a series of image resizing. The scaling of an image does not bring more information about the image, so the quality of the image will inevitably be affected, which also has an impact on the retrieval accuracy. Therefore, it is important to find a suitable scaling ratio to trade-off the retrieval accuracy and the adversarial robustness. The effect of scaling ratio on accuracy can be seen in Figure $\color{red}4b$. Taking the Market1501\cite{market1501} dataset as an example, the original size of the dataset image is [128, 64], and the size is uniformly resized to [256, 128] when fed into the CNN. When using image resizing as passive defense, we observe that the network performance hardly drops and gets a satisfactory defense effect if we resize the image to [110, 50] (Approximately 0.8 times the original image) to corrupt the adversarial noise structure. Our passive defense consists of a series of resizing that reszie the image to [110, 50] then to [220, 100], then to [110, 50] again (finally uniformly to the [256, 128]), so it called circuitous scaling. The effect of image scaling\cite{RRP} (only once resizing) and CS on the adversarial noise can be seen in Figure $\color{red}6$, and it can be observed that the adversarial noise at the contour feature is continuously weakened, and the outline of the pedestrian is more clearer.

\begin{figure}[t]
	\centering
	\setlength{\abovecaptionskip}{0.1cm}       
	\setlength{\belowcaptionskip}{-0.4cm}   
	\includegraphics[width=1\linewidth]{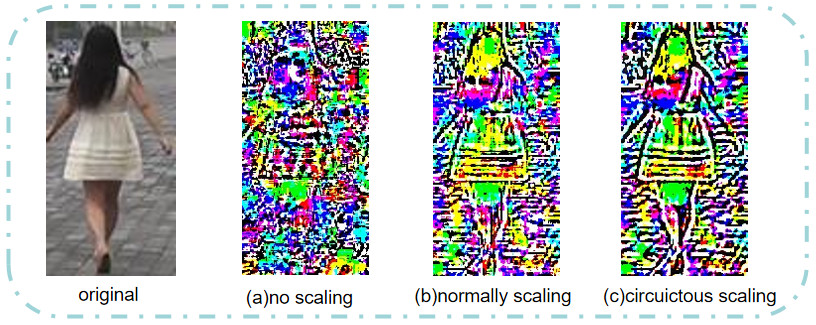}
	\caption{The effect of image scaling and CS on the adversarial noise, where (a) shows the adversarial noise from the original adversarial example, and (b) shows the adversarial noise after reszie the adversarial sample to [110, 50] (then restored to [256, 128]). (c) shows the adversarial noise after CS.}
\end{figure}

\section{Experiments}
In this section, we evaluate our LTA by comparing with SMA\cite{Bouniot} and then evaluate the robustness of our approach using cross-domain tests. Finally, we verify the effectiveness of our JAD under white-box attacks and black-box attack.
\subsection{Attack Evaluation and Cross-Domain Tests}
\textbf{Datasets}. Experiments are conducted on Market1501~\cite{market1501} and DukeMTMC~\cite{duke}. The Market1501 includes 1,501 pedestrians captured by six cameras (five HD cameras and one low-definition camera). The DukeMTMC is a large-scale multi-target, multi-camera tracking dataset, a HD video dataset recorded by 8 synchronous cameras, with more than 2,700 individual pedestrians. The above two datasets are widely used in ReID studies.

\textbf{Evaluation criteria}. Following existing works~\cite{market1501}, Rank-k precision and mean Average Precision (mAP) are adapted as evaluation metrics. Rank-1 denotes the average accuracy of the first return result corresponding to each query image. mAP denotes the mean of average accuracy, the query results are sorted according to the similarity, the closer the correct result is to the top of the list, the higher the score.

\textbf{Implementation details}. The proposed adversarial attack and defense algorithm is development based on on PyTorch framework. In our baseline, ResNet50~\cite{resnet} and DenseNet~\cite{densenet} are used as the backbone network in experiments, and the pre-trained ImageNet parameters are adopted for network initialization. Specifically, the stride of the last convolutional block is set to 2. We adopt the stochastic gradient descent (SGD) optimizer for optimization, and the momentum parameter is set to 0.9. We set the initial learning rate as 0.1. The learning rate is decayed by 0.1 every 40 iteration, with a total of 60 training epochs and a batch size of 32 for normal training on both datasets, and 120 training epochs for our method as well as for adversarial training\cite{SongBai}. 

\textbf{Attack evaluation}. The hyper-parameters are unified for fair comparison, the adversarial boundary is set to 5 pixels, the iteration step size is set to 1, and the number of basic iterations is set to 15. Note that in contrast to an adversarial defense problem, lower precision indicates better attack performance. It can be seen from Table $\color{red}1$ that when only one version of the reference image is used, the success attack rate of LTA* is better than that of SMA. The comparison between LTA* and LTA shows that using diverse versions of reference images has a higher attack success rate than using only one version of reference images. The experimental results fully demonstrate that the attack against color features are more aggressive compared to the same type of SMA attack.

\begin{table}[]  
		\setlength{\abovecaptionskip}{0.1cm}
		\setlength{\belowcaptionskip}{-0.4cm}   
		\centering 
		\caption{Evaluation on Market1501\cite{market1501} under a white-box attack on the query set. Where LTA* means that only one version with local differences is used as a reference image, and LTA generates image versions with different local differences in each basic iteration to conduct the metric attack.}
		\begin{tabular}{lllll}
			\toprule
			Attack     & Rank-1  & Rank-5  & Rank-10 & mAP    \\
			\toprule
			No-attack  & 88.4\% & 95.5\% & 97.1\% & 72.1\% \\
			SMA\cite{Bouniot}        & 15.7\% & 26.4\% & 32.7\% & 11.1\% \\
			LTA*(ours) & 15.7\% & 26.2\% & 32.1\% & 11.0\% \\
			LTA(ours) & \textbf{13.3\%} & \textbf{22.4\%} & \textbf{28.1}\% & \textbf{9.6\%}  \\
			\bottomrule
		\end{tabular}
	\end{table}


\begin{table}[t]\small
	\centering 
	\setlength\tabcolsep{3pt}
	\caption{The performance of different models is evaluated on cross-domain dataset. M→D means that we train the model on Market1501\cite{market1501} and evaluate it on DukeMTMC\cite{duke}.}
	\begin{tabular}{ccccc}
		\toprule
		\multirow{2}{*}{Model} & \multicolumn{2}{c}{M→D}        & \multicolumn{2}{c}{D→M}        \\ \cline{2-5}
		& Rank-1 & mAP & Rank-1 & mAP \\
		\midrule
		Baseline               & 36.1\%          & 18.9\%       & 45.7\%          & 19.6\%       \\
		GOAT\cite{Bouniot}    & 23.6\%          & 11.4\%       & 47.3\%          & 18.5\%       \\
		JAD(w/o LHT)(ours)        & 36.9\%          & 18.4\%       & 47.4\%          & \textbf{19.5\%}       \\
		JAD(ours)               & \textbf{42.5\%}          & \textbf{21.5\% }      & \textbf{47.5\% }         & 19.4\%       \\
		\bottomrule
	\end{tabular}
	\vspace{-0.4cm}
\end{table}
\begin{table}[t]
	\centering
	\setlength\tabcolsep{3pt}
	\caption{The performance of normally trained models (Baseline) and our JAD models on Market1501 and DukeMTMC.}
	\begin{tabular}{ccccc}
		\toprule
		\multirow{2}{*}{Methods} & \multicolumn{2}{c}{Market1501\cite{market1501}} & \multicolumn{2}{c}{DukeMTMC\cite{duke}} \\ \cline{2-5}
		& Rank-1          & mAP           & Rank-1         & mAP          \\
		\midrule
		Baseline      & 88.4\%        & 72.2\%        & 78.7\%        & 62.3\%       \\
		Baseline+RK~\cite{re_ranking}                     & 90.2\%        & 84.7\%        & 83.3\%        & 79.3\%       \\
		JAD(ours)                      & 88.7\%         & 70.3\%        & 77.2\%        & 57.8\%       \\
		JAD+RK(ours)                    & 91.0\%         & 85.0\%        & 82.7\%        & 77.0\%       \\
		\bottomrule
	\end{tabular}
\end{table}

\begin{table}[]\smaller
	\setlength{\belowcaptionskip}{-0.4cm}   
	\setlength\tabcolsep{1pt}
	\caption{The performance of normally trained models and our JAD models under white-box attack of the query.}
	\begin{tabular}{ccccc}
		\toprule
		\multirow{2}{*}{Dataset}    & \multirow{2}{*}{Model} & \multicolumn{3}{c}{Rank-1/mAP(\%)} \\ \cline{3-5}
		&                        & M-IFGSM\cite{SongBai}      & SMA\cite{Bouniot}      & LTA       \\ 
		\midrule
		\multirow{4}{*}{Market1501\cite{market1501}} & Baseline                & 8.1/4.3    & 15.7/11.1 & 13.3/9.6  \\
		& Baseline+RK                   & 13.2/13.0  & 17.6/20.0 & 14.2/16.1 \\
		& JAD(ours)                    & 47.1/27.8  & 79.3/60.1 & 56.5/41.1 \\
		& JAD+RK(ours)                  & 61.3/56.8  & 85.6/80.3 & 66.2/63.9 \\
		\midrule
		\multirow{4}{*}{DukeMTMC\cite{duke}}   & Baseline               & 10.1/5.8   & 15.0/10.4 & 13.0/9.2  \\
		& Baseline+RK                   & 16.8/16.3  & 18.8/19.8 & 15.2/16.3 \\
		& JAD(ours)                    & 30.5/16.3  & 56.7/39.5 & 41.8/27.6 \\
		& JAD+RK(ours)                 & 48.1/43.4  & 69.3/64.5 & 52.4/49.7 \\
		\bottomrule
	\end{tabular}
\end{table}

\textbf{Cross-domain tests}. It is pointed out by\cite{SongBai} that the higher accuracy of the model does not mean that it has better generalization capacity. The defense capabilities of different baselines under the same attack would have been greatly different, and the high accuracy model may even have worse defenses capabilities due to overfit. In response to the above potential problems, we suggest to use cross-domain tests and adversarial defense tests to verify the robustness of the model. Experiments show that the proposed method effectively enhances the generalization capacity of the model, and the Table $\color{red}2$ shows the cross-domain experiments of the proposed method between two datasets, Market-1501\cite{market1501} and DukeMTMC\cite{duke}. We use the state-of-the-art defense model GOAT\cite{Bouniot} for comparison.

In the cross-domain tests of Market1501$\rightarrow$DukeMTMC, it can be seen that JAD (without LHT) enhances the Rank-1 by 3.1 percentage points compared with the baseline, and further enhances by 4.8 percentage points after using LHT. In the cross-domain tests of DukeMTMC$\rightarrow$Market1501, it can be seen that the proposed method (without LHT) enhances the Rank-1 by 3.4 percentage points compared with the baseline, and further enhances by 0.2 percentage points after using LHT. The above shows that the proposed method effectively enhances the generalization capacity of the model, and LHT further enhances the generalization capacity.

\newcommand{ \tabincell}[2]{\begin{tabular}{@{}#1@{}}#2\end{tabular}}
\begin{table}[]\smaller
	\caption{Comparison of baseline, channel fusion (CF), proactive defence (PD), and joint adversarial defense (JAD) in terms of defense accuracy under white-box attack on Market1501\cite{market1501}.}
	\begin{tabular}{ccccc}
		\toprule
		\multirow{2}{*}{\tabincell{c}{Model\\(with RK)}} & \multicolumn{4}{c}{Rank-1/mAP(\%)}            \\ \cline{2-5}
		& No-attack  & M-IFGSM\cite{SongBai}     & SMA\cite{Bouniot}   & LTA       \\
		\midrule
		Baseline               & 90.2/84.7 & 13.2/13.0 & 17.6/20.0 & 14.2/16.1 \\
		CF(ours)                     & 90.8/85.3 & 18.3/16.7 & 18.0/19.6 & 17.1/18.5 \\
		PD(ours)       & 91.5/85.6 & 31.7/28.7 & 58.4/57.1 & 25.2/27.6 \\
		JAD(ours)                     & 91.0/85.0 & 61.3/56.8 & 85.6/80.3 & 66.2/63.9 \\
		\bottomrule
	\end{tabular}
	\vspace{-0.3cm}
\end{table}

\begin{table}[]\smaller
	\setlength{\abovecaptionskip}{0.1cm}
	\setlength{\belowcaptionskip}{0.1cm}   
	\centering
	\setlength\tabcolsep{3pt}
	\caption{Comparison of different resizing combinations in terms of defense accuracy, where $P_1$ means the scaling pattern that resize image to [110, 50] and $P_2$ means resize to [220, 100]. $P_1$→$P_2$ means resize image to [110, 50] then to [220, 100].}
	\begin{tabular}{cccc}
		\toprule
		\multirow{2}{*}{Dateset}    & \multirow{2}{*}{Model} & No-attack       & LTA Attack     \\ \cline{3-4}
		&                        & Rank-1/mAP(\%) & Rank-1/mAP(\%) \\
		\midrule
		\multirow{5}{*}{Market1501\cite{market1501}} & Baseline                      & 90.2/84.7      & 14.2/16.1      \\
		& $P_1$                    & 90.0/84.7      & 25.5/27.5      \\
		& $P_2$                    & 90.3/84.9      & 15.9/17.9      \\
		& $P_1$→$P_2$                 & 90.1/84.8      & 25.8/27.8      \\
		& $P_1$→$P_2$→$P_1$              & 89.8/84.3      & \textbf{31.4/33.1}     \\
		\bottomrule
	\end{tabular}
\end{table}

\begin{table}[]\smaller
	\caption{Comparison of different defense methods in terms of defense accuracy under white-box attacks on Market1501.}
	\begin{tabular}{ccccc}
		\toprule
		\multirow{2}{*}{Model} & \multicolumn{4}{c}{Rank-1/mAP(\%)}                                       \\ \cline{2-5}
		& No-attack  & M-IFGSM\cite{SongBai}    & SMA\cite{Bouniot}    & LTA                \\
		\midrule
		Baseline                 & 90.2/84.7 & 13.2/13.0          & 17.6/20.0          & 14.2/16.1          \\
		AT\cite{SongBai}                     & 86.4/76.9 & 46.8/41.3          & 48.3/47.4          & 49.1/48.4          \\
		AT +CS                 & 86.3/76.1 & 60.6/53.8          & 64.9/61.1          & 62.7/58.5          \\
		JAD(ours)                   & 90.6/84.3 & \textbf{66.9/62.1}          & \textbf{86.5/80.7} & \textbf{73.5/69.5}          \\
		\bottomrule
	\end{tabular}
\end{table}

\begin{table}[]\small
	\setlength{\abovecaptionskip}{0.1cm}
	\setlength\tabcolsep{1pt}
	\caption{Comparison of different defense methods and other baselines in terms of defense accuracy under DMR black-box attack.}
		\begin{tabular}{cccccc}
			\toprule
			\multirow{2}{*}{\tabincell{c}{Model\\(w/o RK)}} & \multicolumn{2}{c}{Rank-1/mAP(\%)} & \multirow{2}{*}{\tabincell{c}{Model\\(with RK)}} & \multicolumn{2}{c}{Rank-1/mAP(\%)} \\ \cline{2-3} \cline{5-6}
			& No-attack        & DMR\cite{Hongjun}       &                        & No-attack       & DMR       \\
			\midrule
			Baseline                 & 88.4/72.2       & 19.8/15.8        & SB\cite{stong_baseline}                       & 95.4/94.2       & 6.2/4.8          \\
			GOAT\cite{Bouniot}                       & 87.5/66.9       & 67.8/46.4       & SB+JAD(ours)                 & 95.1/94.0       & 93.3/91.2        \\
			GOAT+CS                    & 88.0/68.3       & 72.8/50.7        & FR\cite{FastReID}                       & 96.8/95.3       & 24.8/25.9        \\
			JAD(ours)                    & 88.7/70.3       & 81.1/60.7   & FR+JAD(ours)        & 96.3/94.9       & 91.6/90.1    \\
			\bottomrule   
		\end{tabular}
		\vspace{-0.3cm}
	\end{table}

\subsection{Experiments of JAD}
This subsection verifies the effectiveness of the proposed method from white-box\cite{SongBai,Bouniot} attack, black-box\cite{Hongjun} attack and the other baselines\cite{stong_baseline,FastReID}, and shows the effect of each component of the proposed method in defense through the ablation experiment. Then, the state-of-the-art black-box attack DMR\cite{Hongjun} is used to compare defence performance of our JAD and the state-of-the-art defense method GOAT\cite{Bouniot}. Finally, we give a visual analysis of our defense.

We tested our JAD with white-box attacks on Market1501\cite{market1501} and DukeMTMC\cite{duke}, and the attacks include metric-IFGSM (M-IFGSM)\cite{SongBai}, SMA\cite{Bouniot} and the proposed LTA. To be consistent with recent works, we follow the new training/testing protocol to conduct our experiments by k-reciprocal re-ranking (RK)~\cite{re_ranking}. It can be seen from Table $\color{red}4$ that on the two datasets, our JAD has enhanced the Rank-1 in all white-box attacks by more than 40 percentage points after using re-ranking. 

\begin{figure}[t]
	\setlength{\abovecaptionskip}{0.1cm}
	\setlength{\belowcaptionskip}{-0.4cm}   
	\centering
	\includegraphics[width=1\linewidth]{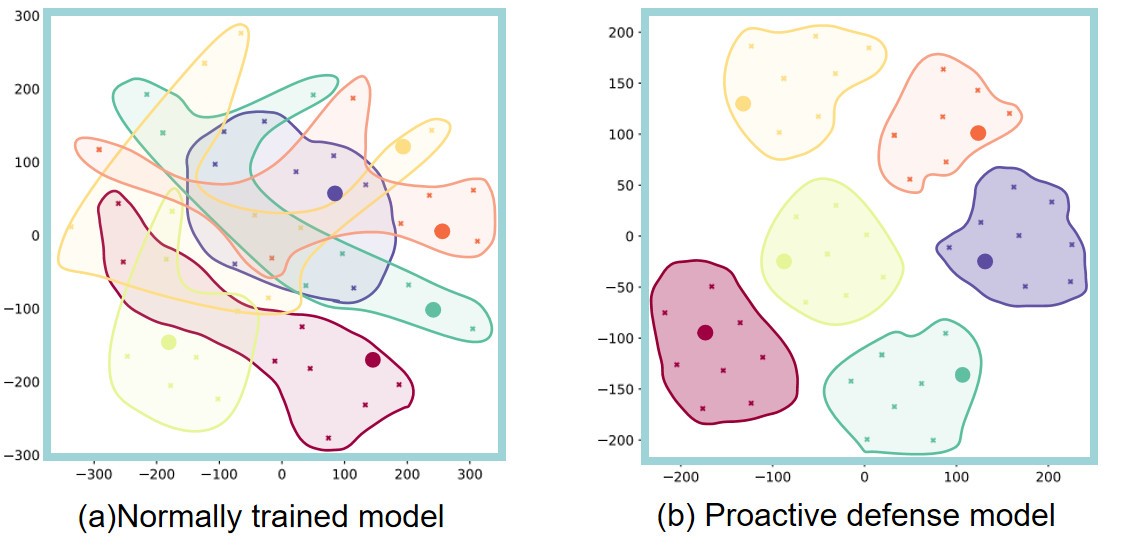}
	\caption{t-SNE~\cite{tsne} visualization of six randomly selected images with different identities on Market1501\cite{market1501}. Each image corresponds to an metric-IFGSM\cite{Bouniot} adversarial example and some randomly generated homogeneous modalities images. The same color means that they are obtained by transformation of the same image. Dots means adversarial examples.}
\end{figure}

$\textbf{Ablation studies}$. We studied the contributions of our channel fusion without augmentation (CF) , proactive defence (PD) and joint adversarial defence (JAD). From Table $\color{red}5$, we can see that defense accuracy increases with the increase of data diversity. As a passive defense, CS further significantly enhances defense performance. Table $\color{red}6$ shows that CS effectively reduces the adversarial effect and thus significantly enhances the defense with negligible performance degradation.

\textbf{Comparison of state-of-the-arts}. AT\cite{SongBai} needs to be customized according to attacks. Specifically, in order to defend an attack, it is necessary to add corresponding adversarial examples to train the model. In Table $\color{red}7$, the original accuracy (no attack) of defense methods based on AT\cite{SongBai} are the average accuracy of the defense models corresponding to the three attacks. AT+CS is the defense model combining the AT\cite{SongBai} and our CS, and it can be seen that the proactive defence and passive defence method CS (that is, our JAD) exert a better combined effect. This shows that our proactive and passive defenses can complement each 	other and have a better gaining effect. As with RRP\cite{RRP}, even if the attacker is aware of the existence of passive defenses, CS can still be effectively defended by a randomization mechanism that allows the resize to fluctuate within a certain range. Compared with AT\cite{SongBai}, our JAD enhances Rank-1 by more than 14.5\% in all white-box attacks.

In the tests of DMR\cite{Hongjun} black-box attack, we use adversarial examples generated by Resnet50~\cite{resnet} to attack the DenseNet~\cite{densenet} model.  GOAT\cite{Bouniot}  model is training based on the adversarial samples generated online by the FNA attack\cite{Bouniot} using triplet loss. GOAT+CS is the defense model combining the GOAT\cite{Bouniot} and our CS. It can be seen from Table $\color{red}8$ that the defense accuracy of our JAD is far better than  GOAT\cite{Bouniot}. In addition, the experimental results show that the JAD is applicable to other baselines~\cite{stong_baseline,FastReID} and performs well. The strong baseline (SB)~\cite{stong_baseline} is implemented based on the Resnet50 backbone network adding the batch normalization neck structure, and FastReID (FR)~\cite{FastReID} is implemented based on the IBN-ResNet101~\cite{ibn} backbone network. The method proposed in this paper has good defense effect in both white-box attack and black-box attack.

\begin{figure}[t]
	\setlength{\abovecaptionskip}{0.1cm}
	\setlength{\belowcaptionskip}{-0.4cm}   
	\centering
	\includegraphics[width=1\linewidth]{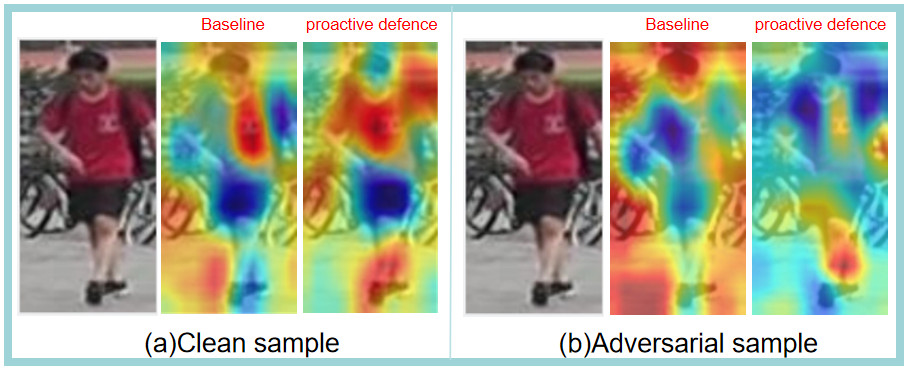}
	\caption{Comparison of Grad-CAM~\cite{Grad-CAM} activation map between normally trained model and our proactive defense model.}
\end{figure}

\textbf{Visualization analysis}. As the show in Figure $\color{red}7$, our proactive defense model with robust to color variations which is insensitive to the variations in the adversarial examples relative to the original examples. Therefore, we can observe that the features of adversarial example and homogeneous examples exhibit clustering effects.

Grad-CAM~\cite{Grad-CAM} uses the gradient information flowing into the last convolutional layer of the CNN to visualize the importance of each neuron in the output layer for the final prediction, by which it is possible to visualize which regions of the image have a significant impact on the prediction of a model. As shown in Figure $\color{red}8b$, we can see that the adversarial example successfully distracts the attention of the normally trained model and activates the opposite parts, while the our proactive defense model is still effectively activating some important parts.

\section{Conclusion}
In this paper, we proposed a color attack method (LTA) based on the local transformation, and further propose a  joint adversarial defense method (JAD) based on the feature-invariance mechanism to enhance the adversarial robustness of ReID. Finally, we use different network structures and baselines under different attack modes to conduct comparative experiments to verify the effectiveness of proposed attack method and joint defense method. Our future goal is to further enhance the stability of the proactive defense model, because we experimentally observed that the limitations of proactive defense are regular. Therefore, we will try to consider cross-datasets when training the model, and update the parameters of the model when a model is improved on the original and cross-domain dataset.

\section*{Acknowledgement}
\noindent
This work was supported by the National Natural Science Foundation of China under Grant Nos. U1805263, 61976053, 61672157. 

{\small
\bibliographystyle{unsrt}
\bibliography{egbib}
}

\end{document}